# Automatic Radish Wilt Detection Using Image Processing Based Techniques and Machine Learning Algorithm

Asif Ashraf Patankar[a], Hyeonjoon Moon[a,*]

[a]Department of Computer Science and Engineering, Sejong University, Seoul 05006, South Korea

**Abstract**

Image processing, computer vision, and pattern recognition have been playing a vital role in diverse agricultural applications, such as species detection, recognition, classification, identification, plant growth stages, plant disease detection, and many more. On the other hand, there is a growing need to capture high resolution images using unmanned aerial vehicles (UAV) and to develop better algorithms in order to find highly accurate and to the point results. In this paper, we propose a segmentation and extraction-based technique to detect fusarium wilt in radish crops. Recent wilt detection algorithms are either based on image processing techniques or conventional machine learning algorithms. However, our methodology is based on a hybrid algorithm, which combines image processing and machine learning. First, the crop image is divided into three segments, which include viz., healthy vegetation, ground and packing material. Based on the HSV decision tree algorithm, all the three segments are segregated from the image. Second, the extracted segments are summed together into an empty canvas of the same resolution as the image and one new image is produced. Third, this new image is compared with the original image, and a final noisy image, which contains traces of wilt is extracted. Finally, a $k$-means algorithm is applied to eliminate the noise and to extract the accurate wilt from it. Moreover, the extracted wilt is mapped on the original image using the contouring method. The proposed combination of algorithms detects the wilt appropriately, which surpasses the traditional practice of separately using the image processing techniques or machine learning.

## 1. Introduction

A disease in plant production significantly reduces the quantity as well as quality of the production yield of the agricultural products. It also consumes a precious amount of cultivation time to manually process and monitor the health of the crops and the diseases in the crops. Presently, the highest concern for agricultural researchers other than feeding a million people and consistently monitoring for the development of weeds in the same suitable climate as the main crop of production is to deal with the spread of disease through the cultivation by infested soil movement.

Firstly, this paper outlines the previous work that was conducted by researchers and the conclusions of their experiments in the said domain. Secondly, this paper proposes a state-of-the-art technology, which applies machine learning to the current improvised methods. Finally, this research compares the results with the current research and challenges. Moreover, the paper focuses on the employment of the proposed research in an industrial computerized system for agricultural use with minimal knowledge of information technology.

The image processing techniques during the past recent years has proven successful with detecting diseases in agricultural crops. The process starts with an analysis of the images that were captured with different environmental conditions. It reduces the manual task time for agricultural researchers and also the traveling time that is required. The experiments can be performed locating remotely. In this cause, a sensor-based technology Unmanned Aerial Vehicle (UAV) can be used.

---

[*] Corresponding author. Tel.: +0-000-000-0000; fax: +0-000-000-0000 .
  *E-mail address:* hmoon@sejong.ac.kr

UAV's are remote-controlled aircrafts that have proven to provide advantageous benefits due to its impact on operational costs and flexibility. Because of the multi functionality use of UAV areas, such as traffic monitoring and hazardous conditions, which include the occurrence of fires and rescuing civilians who are effected by earthquakes, are on a rise [1,2]. The major advantage for an agricultural researcher from a UAV is the spatial resolution imagery feature [3]. On the other hand, the color spectral of the images captured by a UAV is superior to the color impressions perceived by the human eye.

Radishes are perhaps one of the most valuable crops in South Korea, which cover around 10 percent of the total cultivation [4]. There are major efforts conducted to improve the production of the yield of radish crops that are currently implementing the use of technologies in order to achieve the highest yield and precise accuracy to eliminate diseases during the early stage of detection. This motivates the deployment of image processing-based techniques for wilt detection using images from the radish crop fields.

The detection of wilt can be achieved by various pixel color-based methods of image processing and segmentation, which use morphological operations with thresholding. The morphological operations include the opening and the closing operation of the image pixels based on the value of the neighbor pixels. This paper proposes three algorithms in a sequential manner, which include thresholding and morphological methodology, $k$-means clustering, and finally the contouring method. For the detection of wilt in the crop field image, morphological operations is applied after performing thresholding on the image for each category of segments and for more refining of the wilt from the noisy wilt image obtained from the morphological operation, which is applied thereafter by $k$-means clustering [5]. The results from the $k$-means algorithm is mapped on the original image using the contouring method, which shows the crop with wilt. An outline of the proposed algorithm is illustrated in Figure 1, and the steps are explained in the later sections.

Commonly applied color-based techniques for the extraction of textures or features from the image for morphological operations are the vegetative index (VEG), the color index of vegetative extraction (CIVE), and the combined index (COM). In addition, the excess green index (ExG), which is the excess green minus the excess red index (ExGR), and the Otsu method are applied [6–8]. All these methodologies depend on the RGB or the L*A*B* colorspace Hence, their performance degrades with the environmental conditions, such as luminance variations. Before performing the morphological operations, we convert the image from an RGB to an HSV colorspace model. In general, two color spaces are used frequently, which are the HSV and the L*A*B*. Bora et. al. has shown that the performance of the HSV color space model is far better than the L*A*B* [9]. The colors can be defined by the attributes that are listed below.

- The hue, which is the color that is visible to the human eye.
- The saturation, which is the amount of gray in each color from 0 to 100 percent where 0 means gray and 100 means a primary color.
- The value or the lightness, which is the amount of black in each color from 0 to 100 percent where 0 means black and 100 means a color when it is the brightest.

The organization of the proposal is composed as follows. The related work performed by the researchers in the recent past is presented in Section 2. The proposed methodology and the algorithm are explained in detail in Section 3. Section 4 illustrates the experimental results. Finally, Section 5 gives an outline of our conclusion and the future direction of the proposed technique.

## 2. Related work

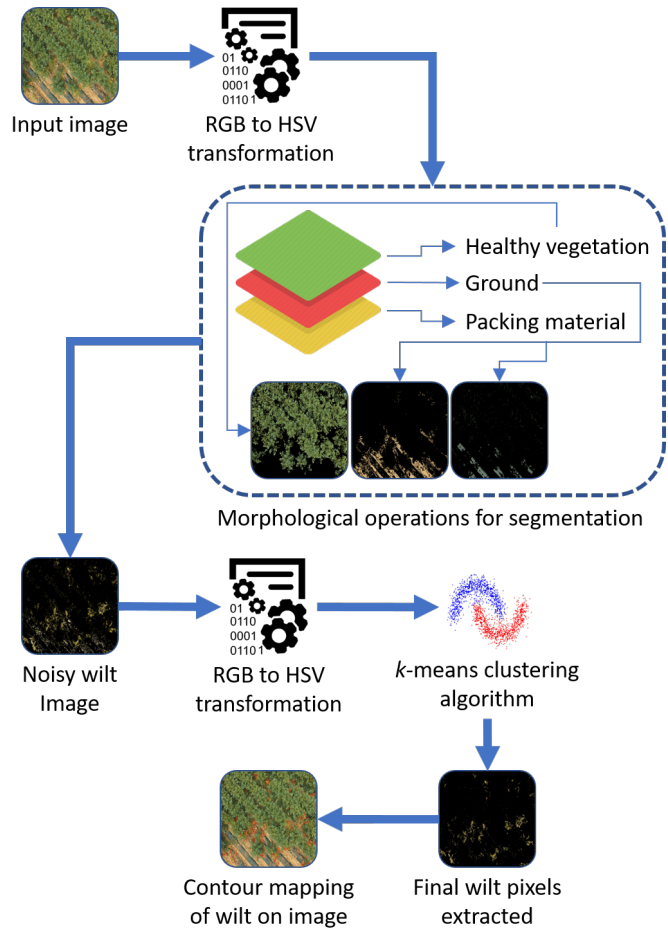

Figure 1. Flowchart of the proposed hybrid algorithm

In the domain of color-based feature extraction a significant amount of work has been done. However, there is not a lot of literature about crop disease detection on HSV colorspace-based feature extraction techniques. The following two researchers have concluded their experiments, which concentrated on an agricultural crop disease and a HSV decision tree, and they are similar to best of the knowledge of the author. Usha et. al. [10] conveys in their proposal how image segmentation and image processing techniques can be implemented to detect plant diseases in the leaves of different fruit trees. They performed and concluded their experiments using MATLAB and color spaces.

Yang et. al. [6] has tested various common color based techniques, which include an excess green index (ExG), an excess green minus an excess red index (ExGR), a vegetative index (VEG), the color index of vegetative extraction (CIVE), a combined index (COM), and also an Otsu algorithm [11]. Their proposed method gave good results compared to the other methods with the HSV decision tree method in order to identify the greenness of maize seedlings in the image. However, the author has concluded that the results may differ if the lightening conditions are poor or strong.

Hamuda et. al. [7] on the other hand has used an HSV decision tree and morphological operations to detect the number of cauliflowers from the video streams, and the eliminated the illuminations in the background at the same time. Their proposed algorithm determined a better performance of detecting 98.91% of the cauliflower plants from video streaming.

The author has further experimented with the results from *k*-means clustering and the extraction of clusters as image frames with contouring algorithms in order to overcome the limitations concluded by the other researchers in their experiments.

In this proposed method, the author has tackled the issues mentioned above and the limitations by further experimenting with the results using a machine learning technique called *k*-means clustering and mapping the extraction of the features with the contouring method afterwards. By comparing the results of this hybrid application of techniques with the others, the results are more consistent and show better accuracy with extracting and mapping the wilt disease in the image with 99.99% accuracy.

## 3. Dataset

The dataset used to demonstrate the proposed research is captured at different intervals of time from the radish fields in Hongchun-gun, Jungsun-gun, and Kangwon-do in South Korea. A commercial Phantom 4, DJI Co. Ltd. Unmanned Aerial Vehicle (UAV) with a 12 mega pixels RGB camera was used for this purpose. The images were taken at varying altitudes from 5 meters to 20 meters from the ground level. The spatial dimensions of the field are 4000 x 3000 pixels with 72 DPI. Figure 2 shows a sample of the original radish field image [4].

## 4. Proposed methodology and the algorithm

The proposed algorithm relates to the initial detection algorithm, which has been explained in the previous research by [7] and [8]. This paper proposes detection methodology to identify wilt in the radish crop fields from the images captured outdoors using a UAV in South Korea.

The experiment results exhibited in this paper are on an image with proportionate exposure to all the segments that were considered for this paper include healthy vegetation and the ground and packaging material. Even though the experiments were performed on 216 radish field images, the images were taken at different altitudes, which included 5 meters (97 images), 7 meters (71 images), 10 meters (39 images), and 20 meters (9 images).

Our pipeline consists of the following five steps that are illustrated below.

- Step 1. The image is converted from an RGB to an HSV colorspace model format.

- Step 2. After converting the image, the morphological operations, such as eroding, dilating, opening, and closing are applied on the image to extract the different categories, which include healthy vegetation, the ground, and the packaging material, of the images using thresholding for each category. The hue, the saturation and the value distribution analysis performed for the proposed experiment can be understood by Figure 3 and Table 1. After the segmentation of the entire image using the above algorithm, the segment images are extracted as shown in Figure 4(a) to Figure 4(d), respectively.

- Step 3. Each of the three categories of images, which include the number of pixels of each category, is then segregated from the original image, and the number of pixels consisting of wilt and other background noises, such as weeds. The paper describes it as a noisy wilt image, which is shown in Figure 5.

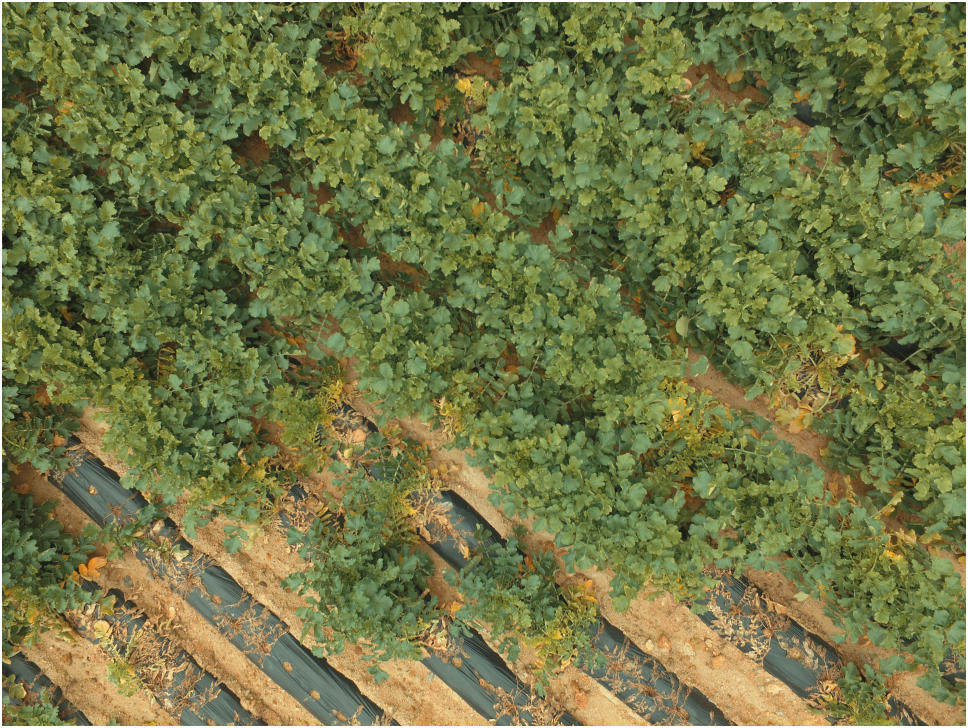

Figure 2. Original radish field image with resolution 4000 x 3000

- Step 4. Later, *k*-means, which is a machine learning algorithm, is applied on the noisy wilt image for the segmentation of the wilt and to separate it from other noises in the image. The image is converted again to an HSV colorspace model format before applying the *k*-means, which is shown in Figure 6. Furthermore, we applied the elbow method to find the optimal number of clusters for the *k*-means algorithm, which is shown in Figure 8 and Table 2.

- Step 5. Finally, a contouring method is used for mapping the wilt (pixels values) on the original image to show the output image that consists of wilt crops.

Table 1. HSV channel distribution analysis for each category

| HSV Channels | Healthy Vegetation | Ground | Packing Material |
|---|---|---|---|
| Hue | 30 Low | 15 Low | 43 Low |
|  | 65 High | 20 High | 179 High |
| Saturation | 59 Low | 85 Low | 0 Low |
|  | 255 High | 255 High | 255 High |
| Value | 43 Low | 35 Low | 0 Low |
|  | 255 High | 255 High | 255 High |

## 4.1. Input image

The method was tested on 216 images captured at different altitudes. We demonstrate our entire algorithm using the image shown in Figure 3. The images were captured with a UAV in distinctive environmental conditions, which included cloudy, shady, partially cloudy, or shady and sunny. Various other factors, such as partial crop appearance in the image, variation in light, different angles of shadows cast on the ground (changing the intensity of light of the ground pixels), and other background factors, such as different boundary soils and stones are also considered. Each image has a frame resolution of 4000 x 3000 pixels.

## 4.2. Image transformation

Initially the image colorspace channels is changed from RGB to HSV channels. Each of the categories, which include healthy vegetation, the ground, and packaging material, are segmented by setting the HSV channels between specific values that determine the minimum and the maximum threshold of each ROIs. It can be seen in the graph in Figure 7 that the hue, the saturation and the value channels of each category differ with each other in probability on the scale of 0 to 255 for the saturation and the value channels and 0 to 179 for the hue channel. Then the morphological operations, which are eroding, dilating, opening, and closing, are applied on the image. In both the algorithms proposed in this paper, which include the morphological operations and the *k*-means, the images are first converted into an HSV color space format. The simplest conversion from RGB to HSV colorspace channels is expressed in Equation (1), Equation (2), and Equation (3).

$$H = \tan\left[\frac{3(G-B)}{(R-G)+(R-B)}\right] \quad (1)$$

$$S = 1 - \frac{\min(R,G,B)}{V} \quad (2)$$

$$V = \frac{R+G+B}{3} \quad (3)$$

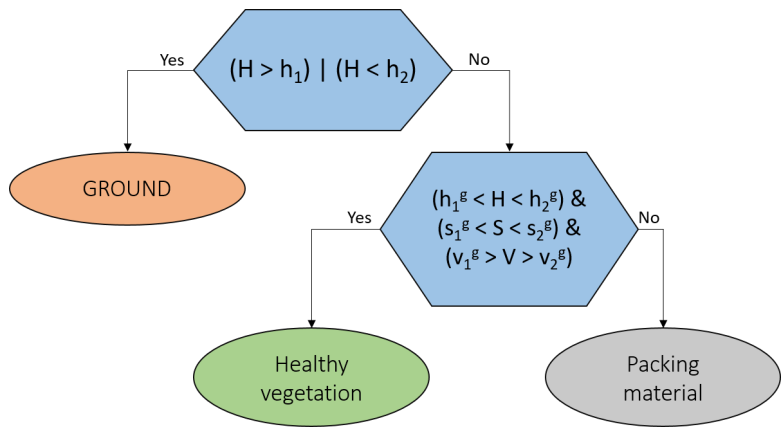

Figure 3. Extraction of three segments thresholding HSV decision tree

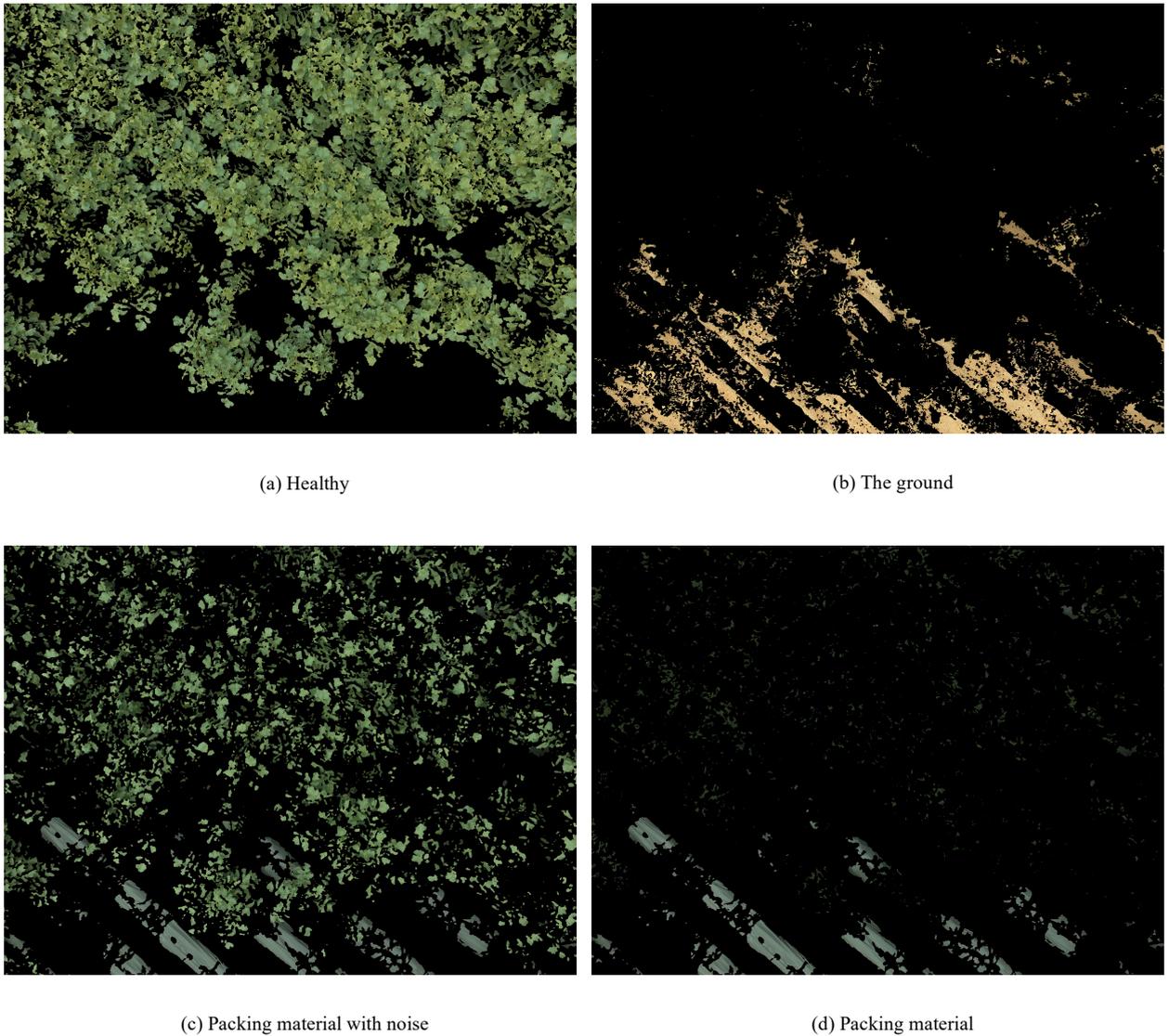

Figure 4. Pixel frame of each segment extracted by applying an HSV decision tree using mathematical morphology operations.

The whole image is classified into four categories, which include healthy vegetation (the green plantation), the ground, the packaging material, and wilt [12]. The original image of the field captured by the drone can be seen in Figure 3.

### 4.3. Morphological operation on the image

The morphological operations, which the name suggests, deal with the non-linear operations in the image processing related to its shape or morphological features. Two of the basic morphological operations, which include erosion and dilation, are applied on the image. The erosion operation followed by the operation is known as the opening operation of an image by a specific structuring element, which eliminates the small noises in the image. The dilation operation followed by the erosion operation is the closing operation of an image by a specific structuring element, which adds a layer of a number of pixels to the inner and the outer boundaries of the regions applied [13].

The algorithm to detect the different categories, which include healthy vegetation, the ground, and the packaging material of ROIs, are extracted as same way as explained by [6]. Yang et. al. [6] has used the hue, the saturation, and the value thresholding for extracting the maize seedlings in the image. Thresholding techniques are generally applied on the original image for transformation of the image for the further determination of the classification. Even though these techniques play a vital role in image segmentation, they may degrade the performance if an optimal threshold value is not used. Let's suppose, if the threshold value is too high, a few important crop regions (crop pixels) may get merged into other crop field regions (other classified categories pixels), which may lead to under-segmentation, but on the contrary if the threshold value is too low, it may lead to over-segmentation.

In this paper, all three categories, which include healthy vegetation, the ground, and the packaging material, are detected and extracted from the image by HSV thresholding, which is explained by [7] and [8,14]. The thresholding is performed at each pixel of the image considering the various environmental conditions of each category. For healthy vegetation, the different color variation in their growth stages is due to shadowing because of the lighting effect and overlapping of the leaves was considered. Secondly, for the ground with lighting effects, the difference in color is due to its dry and wet state. Lastly, for packing material variation in color in few cases. Once the extraction is complete, the pixel values from each category are subtracted from the original image, which leaves behind the pixels consisting of the wilt in the image as shown in Figure 5.

### 4.4. Image with wilt (With noise)

The wilt images extracted after applying the HSV decision tree methodology and thresholding are sometimes observed to contain some noises due to the variation or the direction of light falling on the image or the environmental variations, which can be seen in Figure 5.

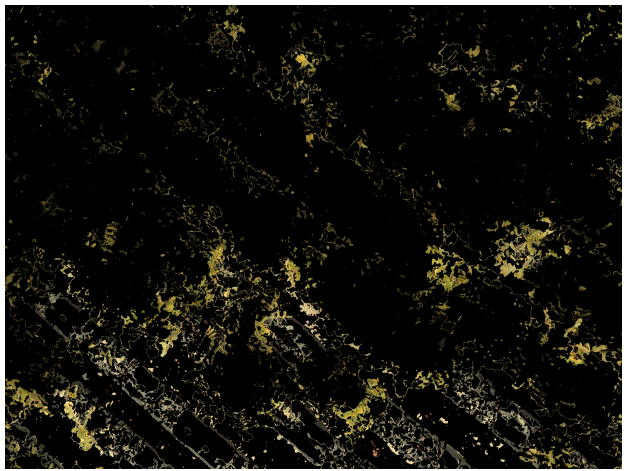 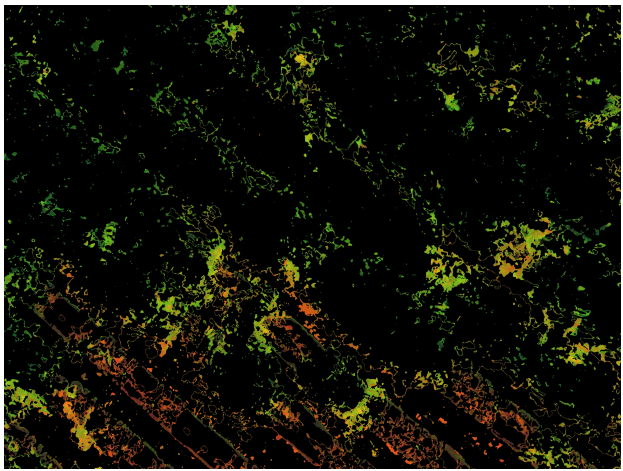

Figure 5. Noisy wilt image                    Figure 5. Noisy wilt image converted into HSV colorspace

### 4.5. k-means clustering applied on the wilt (with noise) image

In this algorithm, the author started iterating with the segmenting of the image using the *k*-means clustering function of OpenCV. OpenCV is a library that is well-known for its application in robotics, HCI, biometrics, and image processing techniques. Other than the *k*-means clustering function and its application in image classification field, which include SVM (Support Vector Machine) and other machine learning application areas, it can also be performed using OpenCV library. Padol et. al. [15] applied the SVM classification and *k*-means clustering to detect diseases in

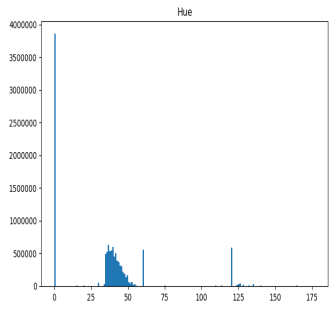
(a) Hue of healthy vegetation

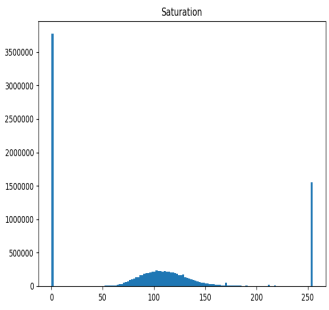
(b) Saturation of healthy vegetation

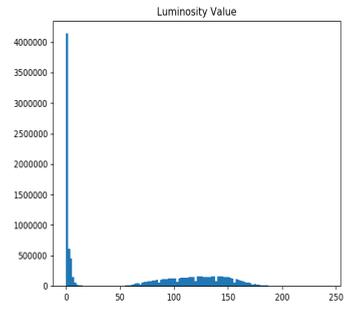
(c) Luminosity Value of healthy vegetation

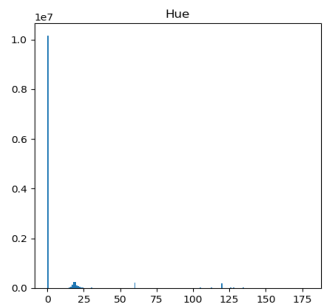
(d) Hue of ground

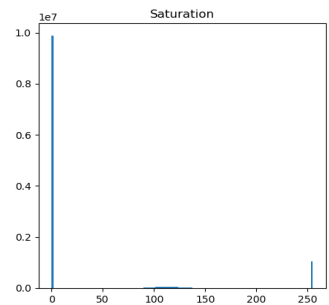
(e) Saturation of ground

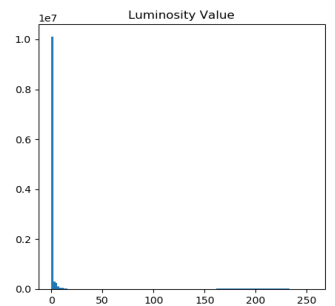
(f) Luminosity Value of ground

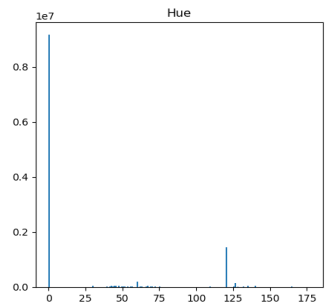
(g) Hue of packing material

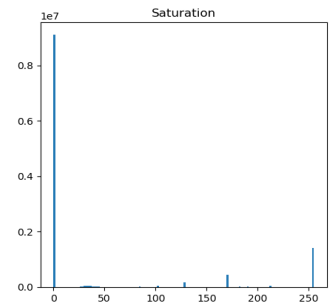
(h) Saturation of packing material

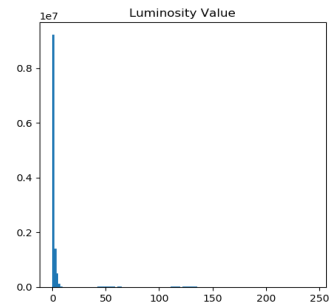
(i) Luminosity Value of packing material

Figure 7. HSV channels of three segments fragmented by thresholding using mathematical morphological operations. Figure 7(a) to Figure 7(c) shows for healthy vegetation, while Figure 7(d) to Figure 7(f) shows for the ground segment and Figure 7(g) to Figure 7(i) shows the packing material H, S, V channel values

grape plants with 88.89% accuracy. Kumari et. al. [16] applied the *k*-means algorithm and a neural network based classifier to detect cotton leaf diseases and with [17,18] using *k*-means as part of the image segmentation. The major advantage of OpenCV is its easy integration with the system hardware [19].

Table 2. Elbow method for extracting a wilt cluster frame

| k clusters | Number of pixels | Visibility of wilt |
|---|---|---|
| 2nd | 3,996,893 | Redundant pixels |
| 3rd | 2,605,116 | Redundant pixels |
| 4th | 2,308,205 | Redundant pixels |
| 5th | 2,362,594 | Redundant pixels |
| 6th | 1,976,609 | Redundant pixels |
| **7th** | **661,417** | **Wilt pixel retrieved** |
| 8th | 626,417 | Dispersed pixels |
| 9th | 624,699 | Dispersed pixels |
| 10th | 606,059 | Dispersed pixels |
| 11th | 539,945 | Dispersed pixels |
| 12th | 527,825 | Dispersed pixels |

$k$-means is a clustering method, which a cluster of a set of pixel values are segregated into $k$ groups called clusters. The algorithm works in two phases. In the initial phase, it calculates $k$ centroids in the pool of pixel values. In the second phase, it segments each different cluster in a pixel-wise arrangement to its nearest centroid forming $k$ clusters mostly using the Euclidian's distance formula. Equation (4) illustrates the distance formula for the $k$-means clustering.

$$dist_{xy} = \sqrt{\sum_{k=1}^{m} (x_{ik} - x_{jk})^2} \qquad (4)$$

OpenCV has a function to perform $k$-means clustering, which takes the parameters as an image, the number of clusters to segment, the number of iterations the clustering should perform, and the labels to extract from an image as per the defined number of clusters. In this proposed method, in order to work with the function, the image is converted to a 32-bit float where the image pixel value range from 0-1.0.

The iterations performed were from 10 to 50 times with the probability of the number of clusters $k$. 20 iterations per image gave the precise output results. For the number of clusters, the different values of $k$ (from 2 to 20) in the $k$-means clustering algorithm were tested for the highest accuracy. The objective was to extract the cluster of the wilt color pixels from the noisy wilt image, which was referred to by the author in the previous section.

Figure 2 shows the extracted number of wilt pixels by changing the number of clusters ($k$) from 2 to 20 in the $k$-means clustering function of OpenCV. The elbow method is used to find the ideal solution for $k$. The number of pixels extracted at each *k* value is written down. At the elbow of the curve where the number of pixels is starting to get saturated is recorded. The optimal solution for extracting an accurate and a consistent number of wilt pixels from the noisy wilt image can be evaluated at *k* = *7* as shown in Figure 8.

Rather than applying the *k*-means clustering on the entire original image and calculating the pixel-level clustering accuracy (PSA) as applied by other researchers, we extracted the pixels from the different clusters and printed it on different canvas using arithmetic looping to differentiate and calculate the wilt pixels from other noisy pixels in the image as shown in Figure 9.

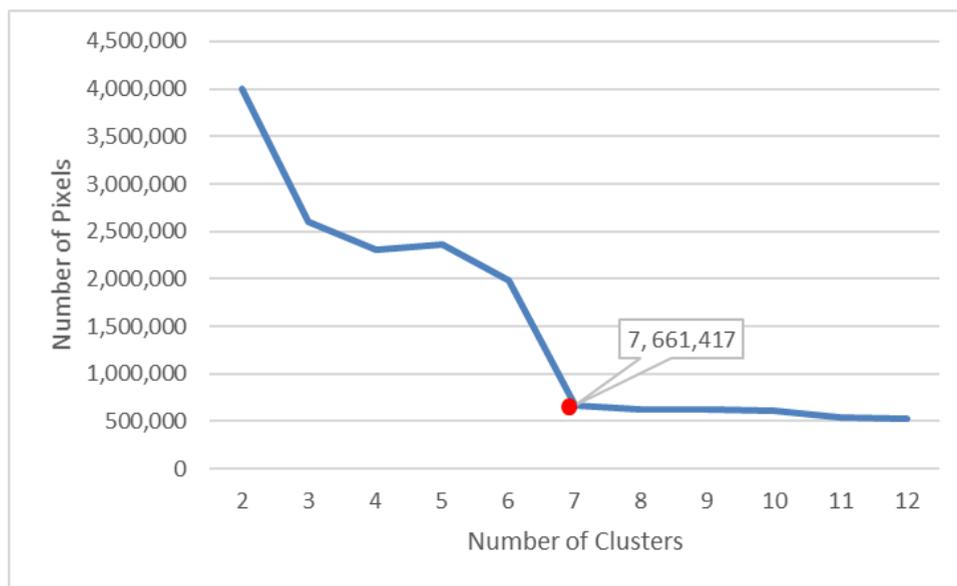

Figure 8. The elbow method for the optimal number of clusters k. The red point shows the optimal solution.

*4.6. Final wilt image*

The filtered wilt from other background noises is retrieved after applying the *k*-means clustering as in Figure 10. The wilt extracted from the *k*-means clustering is pasted on a canvas image with a background that has the same dimensions as 4000 x 3000 color as black [RGB color (0, 0, 0)]. Furthermore, these images with wilt pixels and black color pixels are applied with contouring, which is explained in section 4.7.

*4.7. Contouring*

Contouring has always played an important role in image segmentation and classification in the image processing and computer vision fields. It is a powerful tool that is used to segment the image based on pixels, edges, or regions. In this proposed research, contouring is used for segmenting the already thresholded segment of wilt by *k*-means clustering [20]. A function of OpenCV, which is *findContours*, is used to obtain the ROI of the wilt in the image. The image is treated as a binary 8-bit single-channel image by the *findContours* function, because the non-zero pixels in the image is treated as 1's and the zero pixels are treated as 0's. Also, the black color pixels in the background, gets eliminated, and the wilt pixels are retrieved precisely for contouring.

Finally, a function called *drawContours* from OpenCV is used to plot the fusarium wilt pixels on the original image. This function takes parameters as image, which the wilt is to be mapped and the pixel values are retrieved from the *findContour*s function. Figure 11 shows the mapped wilt pixels on the original image [21].

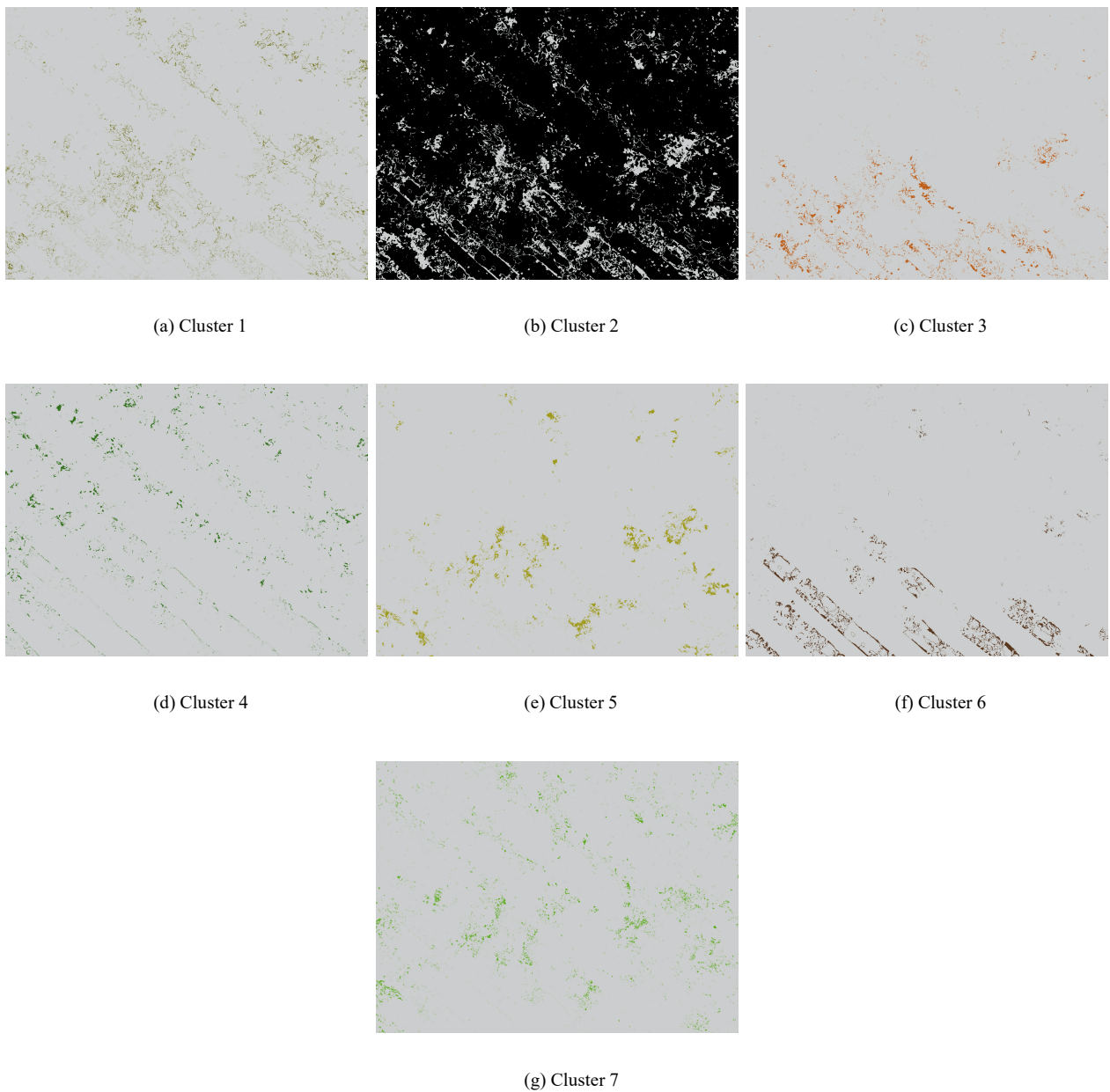

(a) Cluster 1  (b) Cluster 2  (c) Cluster 3

(d) Cluster 4  (e) Cluster 5  (f) Cluster 6

(g) Cluster 7

Figure 9. *k*-means clustering extracted frames out of the noisy wilt image from 9(a) to 9(g)

## 5. Discussion and results

A comparison of the results is performed with the different color spaces and the *k*-means algorithm for the probability analysis. The color spaces included from grayscale, L*A*B*, and YCrCb to HSV. The HSV color space with its minimum and maximum thresholding values followed by the mathematical operations viz. opening and closing operations gave the desired result for the experiment on the radish crop image. Later, *k*-means was applied by observing the number of pixels extracted in each set of clusters, which started from $k = 2$ to $k = 20$. The $k$-means $k = 7$ gave the accurate pixels of wilt in the image.

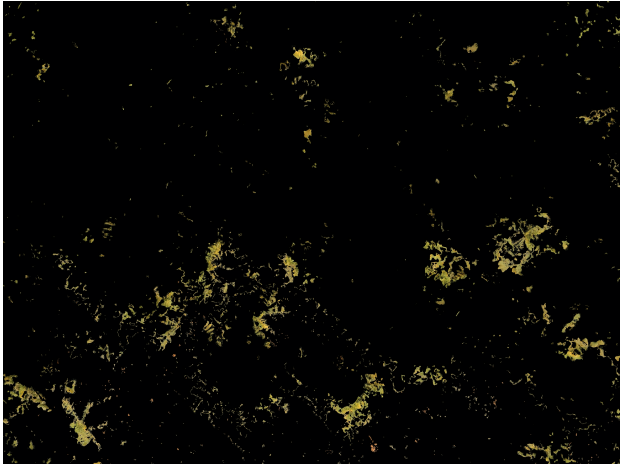 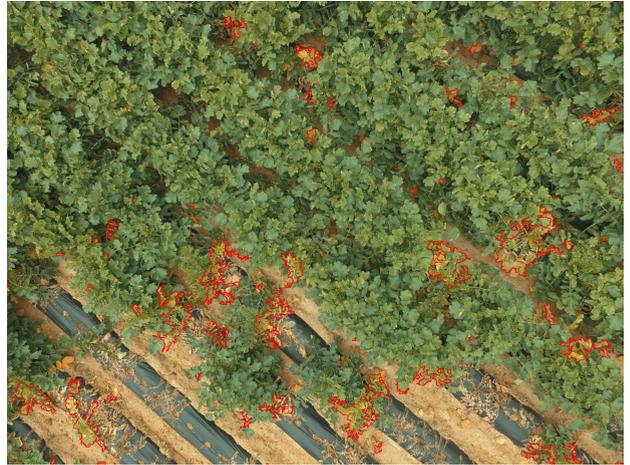

Figure 10. Final wilt image                    Figure 11. Original image with wilt disease mapped on it

As the wilt is just a color pixel in the image, the number of pixels neighboring together must be set to map the wilt on the original image, which is shown in Figure 11. For this operation, the contouring method is used. It can be explained by supposing the area of the number of pixels neighboring each other is more than 10,000 – 15,000 pixels, then it should be mapped in the image.

The number of wilt plants detected in the crop image is 99.99% as tested on the image dataset captured in different lightening conditions. The accuracy is calculated with the number of wilt plants per image feed to the computerized application system for monitoring.

## 6. Conclusion

This paper proposes an image segmentation approach, which can segment the different color elements present in an image with the same hue channel value but with a varied intensity channel value using an HSV decision tree. Initially, the radish field image is converted from an RGB to an HSV colorspace model. The field is classified into three categories that include healthy vegetation, the packaging material, and the ground. Then the pixels of all three, which include the healthy vegetation, the packaging material, and the ground, are extracted according to their hue, saturation, and intensity values compared to each other. The pixel traces left behind in the image after the extraction of the categories mentioned above are the traces of the pixels of wilt in the field. Finally, to eliminate the noises in the wilt image, a machine learning algorithm, which is $k$-means, is applied on the image. The clustered image after applying the $k$-means segmentation shows that the proposed method can recognize the wilt in the image accurately. The limitations although is with determining the HSV channel thresholding values for each fragmented segment as performed for this proposal.

The proposed hybrid algorithm approach can be applied to any image of a vegetation field to extract wilt disease from the image after the appropriate modifications. It can be further extended to locate the wilt in the image by using the coordinates recorded in the information system of the image while capturing the image by UAV.